\documentclass[letterpaper, 10 pt, conference]{IEEEtran}  

\IEEEoverridecommandlockouts                              

\usepackage{cite}
\usepackage{comment}
\usepackage{amsmath, amssymb, amsfonts}
\usepackage{algorithm}
\usepackage{algpseudocode}
\usepackage{graphicx}
\usepackage{textcomp}
\usepackage{stfloats}
\usepackage{xcolor}

\def\BibTeX{{\rm B\kern-.05em{\sc i\kern-.025em b}\kern-.08em
    T\kern-.1667em\lower.7ex\hbox{E}\kern-.125emX}}
\begin{document}
\title{A Resilient Framework for 5G-Edge-Connected UAVs based on Switching Edge-MPC and Onboard-PID Control}
\author{Gerasimos Damigos$^{1*}$, Achilleas Santi Seisa$^{2*}$, Sumeet Gajanan Satpute$^{2}$, Tore Lindgren$^{1}$ and George \\ Nikolakopoulos$^{2}$%
\thanks{This project has received funding from the European Union’s Horizon 2020 research and innovation programme under the Marie Skłodowska-Curie grant agreement No 953454.}
\thanks{$^{1}$ The authors are with Ericsson Research, Lule\aa\,\,}
\thanks{$^{2}$ The authors are with the Robotics and AI Group, Department of Computer, Electrical and Space Engineering, Lule\aa\,\, University of Technology, Lule\aa\,\,}
\thanks{$^{*}$ The authors contributed equally}
\thanks{Corresponding Authors' email: {\tt\small (achsei, geonik)@ltu.se, gerasimos.damigos@ericsson.com}}
}
\maketitle
\begin{abstract}
In recent years, the need for resources for handling processes with high computational complexity for mobile robots is becoming increasingly urgent. More specifically, robots need to autonomously operate in a robust and continuous manner, while keeping high performance, a need that led to the utilization of edge computing to offload many computationally demanding and time-critical robotic procedures. However, safe mechanisms should be implemented to handle situations when it is not possible to use the offloaded procedures, such as if the communication is challenged or the edge cluster is not available. To this end, this article presents a switching strategy for safety, redundancy, and optimized behavior through an edge computing-based Model Predictive Controller (MPC) and a low-level onboard-PID controller for edge-connected Unmanned Aerial Vehicles (UAVs). The switching strategy is based on the communication Key Performance Indicators (KPIs) over 5G to decide whether the UAV should be controlled by the edge-based or have a safe fallback based on the onboard controller.

\end{abstract}
\begin{IEEEkeywords}
Edge Robotics; 5G; UAV; MPC; Resiliency.
\end{IEEEkeywords}

\section{Introduction}
\label{intro}
Cloud and edge computing have emerged in the field of robotics, and the terms of cloud robotics~\cite{wan2016cloud, hu2012cloud, saha2018comprehensive} and edge robotics~\cite{groshev2022edge, haidegger2019robotics, 9837289} are becoming a trend in the scientific world. At the same time, 5G is providing an ideal communication environment thanks to the increased performance and the additional available features like the Quality of Service (QoS)~\cite{damigos2023performance}. However, the safe utilization of remote cloud or edge computing resources requires the consideration of onboard safety fallback actions for mission-critical applications. Furthermore, communication Key Performance Indicators (KPIs) can provide useful information on the status of the system, that can be used to trigger a series of safety actions.

\label{motivation}
This article deals with the challenge of communication uncertainty within cloud or edge-connected robots over 5G. Even though edge computing and 5G networks can provide minimal latency, robust networking, and reliable access to external computational resources, still the need for onboard processing for safety reasons is essential. The edge-based algorithms can provide optimized behavior for the system, while the onboard backup actions can provide redundancy in case of degraded communication. Communication degradation can happen in cases such as out-of-coverage scenarios or overloaded network cells. In the authors' previous works~\cite{9831701, seisa2022cnmpc}, edge-based architectures were presented for offloading time-critical applications, while these contributions were demonstrated with edge-based Model Predictive Control (MPC) schemes over 5G networks~\cite{damigos2023towards}. The need for edge resources for the execution of the edge-MPC was verified with a series of experiments in~\cite{robotics11040080}. The previous works were focused on the Kubernetes (K8s) architecture, which has also been used for this article, and how the MPC could be offloaded to the edge in an optimal manner. However, no actions were considered in case of communication issues.

\begin{figure}[ht!]
	\centering
	\includegraphics[width=0.9\columnwidth]{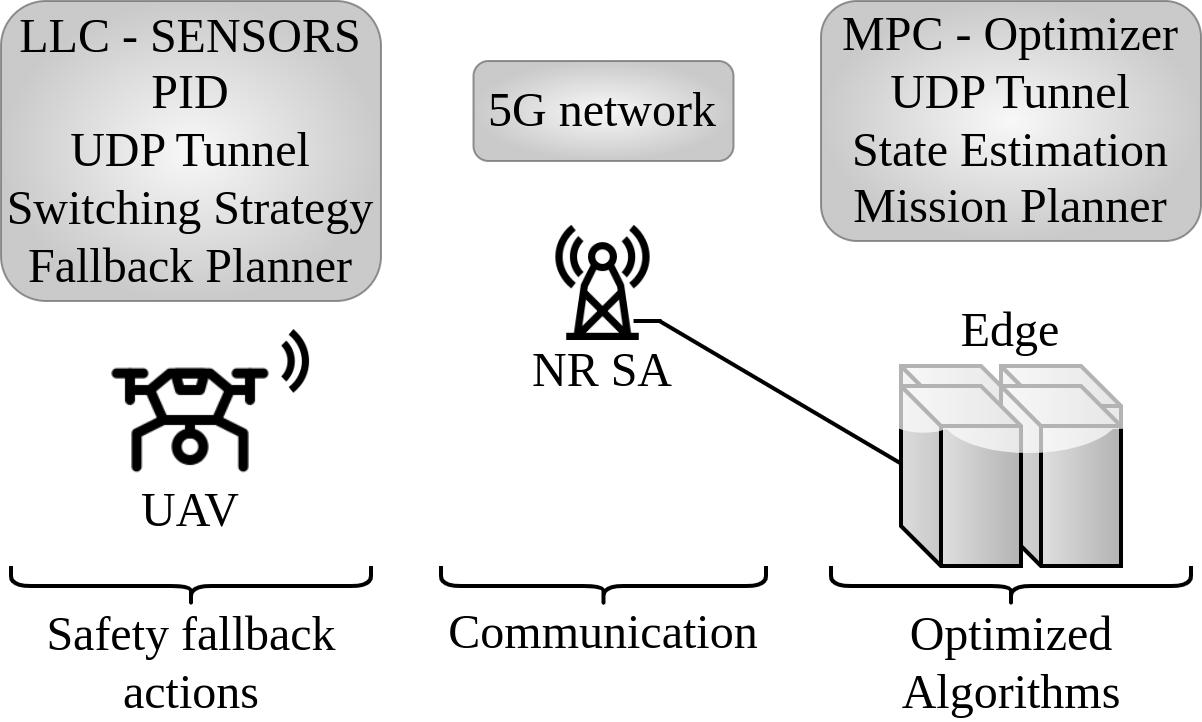}
  	\caption{System Overview. The safety fallback actions include the switching strategy on the UAV along with the fallback planner and the PID control. The optimized algorithms on the edge include the MPC, the optimizer, and the state estimation. Additionally, a 5G network has been established for communication}
  	\label{fig:concept}
\end{figure}

In the novel proposed framework, shown in Fig.~\ref{fig:concept}, a switching strategy is developed and is responsible to decide the input of the Low-Level Controller (LLC) of the Unmanned Aerial Vehicle (UAV). The modules shown in Fig.~\ref{fig:concept} run either offboard (Edge), and are mainly focused on optimizing the behavior of the system, or onboard (UAV onboard computer) and are mostly focused on the UAV operation and safety actions. In this framework, the switching strategy, which is running onboard, can utilize either the edge-MPC or the onboard control commands, based upon the status of the communication link. The chosen onboard controller is a Proportional–Integral–Derivative (PID) controller because it is computationally light and can run on any UAV's onboard computer. To establish robust communication between the UAV and the edge, 5G networks have been used, while for the development of the source code and the messaging between the different modules, the Robotic Operating System (ROS) framework was utilized as in~\cite{toffetti2020cloud}.

\label{soa}
While some studies deal with communication issues with edge algorithms that can tolerate latency~\cite{papadimitriou2022multi, 8473376, viswa}, switching mechanisms are essential for system redundancy. Even though switching from cloud/edge servers to local operation is a critical functionality for autonomous systems, there are not many works that are addressing this problem. As such, some articles are proposing a switching mechanism between the cloud and the edge operation. In~\cite{7749210}, autonomous vehicles are controlled by the cloud, since multi-sensor data for multiple autonomous vehicles can be handled better by the cloud. However, when latency measurements are higher than a threshold, the switching method activates the edge controller. Other works, like~\cite{8549396}, promote switching between different edge servers to handle communication issues for autonomous vehicles. These studies, though, do not consider cases when the latency is very high or when the communication is completely lost, thus making the need for onboard control crucial.

A self-reliant MPC and a replacement controller have been proposed in~\cite{9074603}. The self-reliant MPC is operating on the cloud, while the replacement controller is providing improved control performance when there are relatively long duty-standby transitions that lead to degraded performance of the self-reliant MPC. In~\cite{9936595} and~\cite{9211472}, edge-based and local controllers have been introduced for industrial control systems. The two controllers cooperate to enhance the performance based on a switching logic. In~\cite{9936595} the local controller outputs the edge controller's commands when it is delivered on time, and it outputs its own command otherwise. The switching logic in~\cite{9211472} is aiming to both guarantee stability and optimal control. The edge controller provides optimal behavior when the system is operating in a performance region, but once the system exits the performance region, the system switches to the local controller to ensure stability, Many approaches have also considered switching schemes from a classical time delay approach as in~\cite{nikolakopoulos2008experimental} but this article is focusing only on edge oriented architectures. 

The authors in~\cite{9683307} proposed a mechanism where a cloud MPC can control independently a system, support a local controller or have the local controller operate independently when the cloud MPC fails, while in~\cite{skarin2020cloud}, a cloud MPC, a local MPC and a Linear Quadratic Regulator (LQR) are utilized to control the system. The cloud MPC is running at a high rate and is responsible for optimally controlling the system. This occurs when the control command originating from the cloud MPC can be available to the system within a decided sampling period. Otherwise, the local MPC, which is running at a lower rate, takes over. Since the computational power locally is limited, the local MPC might not be able to solve the optimization problem and generate control commands. In this case, the local LQR controls the system.

In comparison to all the previous works, our system does not switch from the remote controller to the local based on latency metrics, but it uses the communication KPIs and estimates the position error (the difference between the actual position and the desired position of the UAV) produced by these metrics. By doing so, the sensitivity of the system to time delays is taken into consideration. Thus, the system can reactively switch to the local safe mode, only when the communication is heading the system to undesirable states.

\label{contributions}
The main contribution of this work is the development of a novel resilient strategy that can reactively switch from offboard to onboard controllers based on the availability of a fast and reliable communication link. The switching strategy optimizes the system's behavior by utilizing the advanced edge-based algorithms when the system respects the communication requirements and switches to the onboard safety mode when communication is degraded and considered unstable as relying on offloaded procedures can lead to huge position errors and destabilize the system under such conditions. Furthermore, the switching mechanism provides redundancy when the connection is poor or lost since the function is based on connectivity conditions like signal strength, packet loss, or based on end-to-end KPIs, like latency. These metrics are used by the strategy to estimate a position error, and thus, by switching between onboard and edge-based controllers, it can be used to keep the position error bounded. The switching, along with additional computational light components that were developed on the UAV's onboard computer, acts as a safety resilient layer for the overall system. Finally, the whole system goes through a series of experiments for the evaluation of the switching strategy.

The rest of the article is organized into three sections. Section~\ref{system_overview} is describing the overall system, the modules, and the components of the system. The main modules and components are introduced in subsections and the switching strategy is analyzed. In Section~\ref{experiments}, the experimental setup is presented, among with the results and the evaluation of the system. Finally, Section~\ref{conclusions} concludes with the justification of the article's conceptions, and some interesting future directions and implementations are proposed.

\section{System Overview and Safety Actions}
\label{system_overview}
In this work, we are focused on developing resilient fallback actions for 5G-edge-enabled UAVs based on latency, application layer dropped packets, and signal strength measurements. Depending on these communication KPIs, the system decides through a switching mechanism, whether the UAVs should be controlled by the edge algorithms, which are developed utilizing the Kubernetes PODs technology for optimized behavior (offboard control mode) or whether the onboard modules should take over (onboard control mode). To ensure the functionality of the switching mechanism, many components were utilized and developed. Moreover, a UDP tunneling has been developed for this work in order to forward the ROS messages from the onboard computer to the PODs of the Kubernetes server.

\subsection{Edge-MPC and Onboard-PID}
\label{controllers}
The proposed mechanism is switching between an MPC, which is running on the edge, and a PID, which is running on the UAV's onboard computer. It receives as an input the control command signal, from both the MPC and the PID, and outputs one of the two signals. This output will be the input of the UAV's LLC.

\begin{figure*}[ht!]
	\centering
	\includegraphics[width=0.8\textwidth]{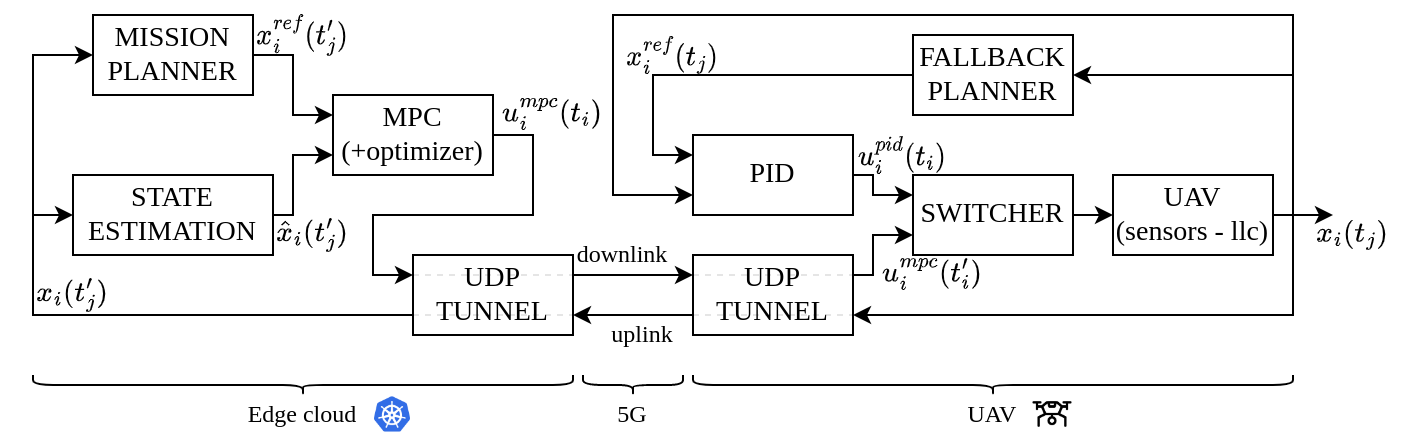}
  	\caption{Control block diagram of the system}
  	\label{fig:block_diagram}
\end{figure*}

\subsubsection{Edge Model Predictive Control}
\label{MPC}
A 5G-enabled UAV that uses an edge server, as an external computational unit, experiences minimal latency in the uplink and downlink communication direction. Fig.~\ref{fig:block_diagram} depicts the observed latency in the uplink and downlink direction. For both the onboard controller and the remote controller on the edge server, the UAV is described as a robot with six degrees of freedom and a fixed body frame, as presented in~\cite{viswa} and described in Eq.~\ref{eq:kinematics}.

\begin{align}
&\dot{p}(t) = v_{z}(t) \nonumber\\
&\dot{v}(t) = R_{x,y}(\theta,\phi) \begin{bmatrix} 0\\ 0\\ T_{ref}\end{bmatrix} + \begin{bmatrix} 0\\ 0\\ -g\end{bmatrix} - \begin{bmatrix} A_{x} & 0 & 0\\ 0 & A_{y} & 0\\ 0 & 0 & A_{z}\end{bmatrix}v(t) \label{eq:kinematics}\\
&\dot{\phi}(t) = \frac{1}{\tau_{\phi}} (K_{\phi} \phi_{ref}(t) - \phi(t)) \nonumber\\
&\dot{\theta}(t) = \frac{1}{\tau_{\theta}} (K_{\theta} \theta_{ref}(t) - \theta(t)) \nonumber
\end{align} 

The position of the UAV is denoted as $p = [p_{x}, p_{y}, p_{z}]^{T}$ and the linear velocity is denoted as $v = [v_{x}, v_{y}, v_{z}]^{T}$. A rotation matrix that describes the attitude of the UAV in Euler form is denoted by $R(\theta(t), \phi(t)) \in SO(3)$, where $SO(3)$ is the $3D$ rotation group and the roll and pitch angles are denoted respectively as $\phi \in [-\pi, \pi]$ and $\theta \in [-\pi, \pi]$. The variables with the subscript `\textit{ref}' represents the desired value. The only parameters that affect the acceleration are the magnitude and the angle of the thrust vector produced by the motors, the linear damping terms $A_{x}, A_{y}, A_{z} \in R$ and the gravity of earth $g$. This can be derived from Eq.~\ref{eq:kinematics}. A first-order system is used to model the relationship between the attitude (roll/pitch), and the referenced terms $\phi_{ref}$ and $\theta_{ref} \in R$, with gains $K_{\phi}$ and $K_{\theta} \in R$ and time constants $\tau_{\phi}$ and $\tau_{\theta} \in R$. Additionally, a Lower-Level attitude Controller (LLC) takes as input the thrust, roll, and pitch commands and generates the motor commands for the UAV. The control command values are saturated, as will be described in Section~\ref{pid}. Note here that the position and the linear velocity in the described setup are obtained through the sensing system for the UAV and sent to the edge server over the uplink channel.

For the cost function, a related optimizer is assigned to find an optimal set of control actions, defined by the cost minimum of the cost function $J$ described from Eq.~\ref{eq:cost_funtion}.

\begin{align}
&J = \sum_{j=1}^{N} \underbrace{(x_{d} - x_{k+j|k})^{T} Q_{x} (x_{d} - x_{k+j|k})}_{state \quad cost} \nonumber\\
&+ \underbrace{(u_{d} - u_{k+j|k})^{T} Q_{u} (u_{d} - u_{k+j|k})}_{input \quad cost} \label{eq:cost_funtion}\\
&+ \underbrace{u_{k+j|k} - u_{k+j-1|k})^{T} Q_{\delta u} (u_{k+j|k} - u_{k+j-1|k})}_{control \quad actions \quad smoothness \quad cost} \nonumber
\end{align} 

where $N$ is the prediction horizon of the MPC, $x = [p, v, \phi, \theta]^{T}$ is the UAV's state vector and $u = [T, \phi_{d}, \theta_{d}]^{T}$ is the control input. $Q_{x} \in \mathbb{R}^{8x8}$ is the matrix for the state weights, $Q_{u}$ is the matrix for the input weights, and $Q_{\delta u} \in \mathbb{R}^{3x3}$ is the matrix for the input rate weights.

\subsubsection{State Estimation}
\label{state_estimation}
The proposed switching strategy consists of multiple modules and components. The state estimation module is designed to account for the data that flow from the UAV to the edge server, i.e., the uplink, while the error estimation module on the UAV accounts for the data that flow from the edge server to the UAV, i.e., the downlink. The more complicated case of the uplink direction is handled by estimating the actual state of the UAV by utilizing the received delayed (from the uplink link) state on the edge server. This method is inspired by the work described in~\cite{viswa}.

As depicted in Fig.~\ref{fig:block_diagram}, the captured state of the robot is delayed by $l_{u} = t'_{j} - t_{j}$. In order to compensate for the $l_{u}$, the state of the robot is estimated when the data arrive at the edge server. A timestamp field in the robot's state is used to calculate $l_{u}$ and the model of the system to predict the actual state. Let $p(t)$ be the UAV's position, and $v(t) = \dot{p}_(t)$ be the UAV's velocity, then the estimated position and velocity are formulated by Eq.~\ref{eq:pos2vel}.

\begin{align}
    & \hat{p}(t) = p(t - l_{u})  \nonumber \\
    & \implies \hat{v}(t) = v(t - l_{u}) \label{eq:pos2vel} 
\end{align}

To track the future state, the uplink delay $l_{u}$, has to be taken into account, thus, the latter expression for the velocity is expressed by Eq.~\ref{eq:vel_est_fut_1}.

\begin{align}
    \hat{v}(t + l_{u}) &= \hat{v}(t) + \int_{t-l_{u}}^{t} \dot{v}(t) dt \label{eq:vel_est_fut_1} 
\end{align}

The integral term in \eqref{eq:vel_est_fut_1} is simplified using a Taylor series approximation and ignoring the higher order terms (since $l_{u} ^2 << l_{u}$) as denoted in Eq.~\ref{eq:vel_est_fut_2}.
\begin{align}
    \hat{v}(t + l_{u}) &= \hat{v}(t) + \dot{v}(t) l_{u} \label{eq:vel_est_fut_2}.
\end{align}

Respectively, the expression for the estimated position is described by Eq.~\ref{eq:pos}.

\begin{align}
    \hat{p}(t + l_{u}) &= \hat{p}(t) + v(t) l_{u} \label{eq:pos}.
\end{align}

Finally, the delayed values regarding the roll and pitch can be derived in a similar manner.

\subsubsection{Onboard PID Control}
\label{pid}
In the onboard control mode, the PID controller takes over based on the fallback switching signal and generates the corresponding control actions to continue the UAV mission. These control actions are produced by less advanced algorithms in comparison to the ones that are produced from the edge controller, but their generation requires much less computational effort. The controller is a standard PID controller with gains $K_{P}, K_{I}, K_{D}$ for the proportional, integral, and derivative terms respectively, that take as input the odometry data, $x_{i}$ (position of the UAV $p_{i}$) from the UAV sensors and generates roll, pitch, yaw, and thrust commands, $u^{PID}_{i}$. These commands are saturated to an upper $u^{th}$ and lower $-u^{th}$ value as expressed by Eq.~\ref{eq:saturation} so the LLC will not receive extreme control commands.

\begin{align}
    & u_{i} = u^{th}, \hspace{0.5cm} if \hspace{0.2cm} u^{PID}_{i} \geq u^{th} \nonumber \\
    & u_{i} = -u^{th}, \hspace{0.23cm} if \hspace{0.2cm} u^{PID}_{i} \leq -u^{th}     \label{eq:saturation} \\
    & u_{i} = u^{PID}_{i}, \hspace{0.2cm} elsewhere \nonumber
\end{align}
%

\subsection{Switching Strategy}
\label{switching_strategy}
To ensure the UAV's autonomy, the switching strategy, which can also be utilized as a resilient fallback mechanism, is deployed onboard the UAV. Unlike the most common approaches in the literature, which usually employ mechanisms based on the round trip time (RTT) delay~\cite{nikolakopoulos2013switching}, the presented framework utilizes a switching mechanism that is triggered on the estimated error, i.e., how much the acquired trajectory will deviate from the reference trajectory, based on the downlink latency and the measured dropped packets in the application layer. Finally, a radio signaling KPI is employed to account for the non-linear relationship between signal coverage and various latency KPIs.       

\subsubsection{Error Estimation based on the Downlink Latency}
\label{error_estimation}
Due to the downlink latency, we can assume that the UAV's actual position $p_{i}$ may deviate from the reference position $p_{i}^{ref}$, given that the aerial robot has a linear velocity $v_{i}$ different from zero. The estimated error $\hat{e}$, based on the downlink latency ($l_{c}$ and $l_{d}$), does not reflect the overall position error (difference between the actual position and the desired position of the UAV), but it provides an estimation on how the error can vary due to the latency. Of course, there are other parameters that affect the error, such as the uncertainty, the control design defect, the disturbances, etc., but these are not considered in the design of the switching strategy, since the error from these parameters will occur whether we use onboard or offboard controllers.

\begin{figure}[ht!]
	\centering
	\includegraphics[width=\columnwidth]{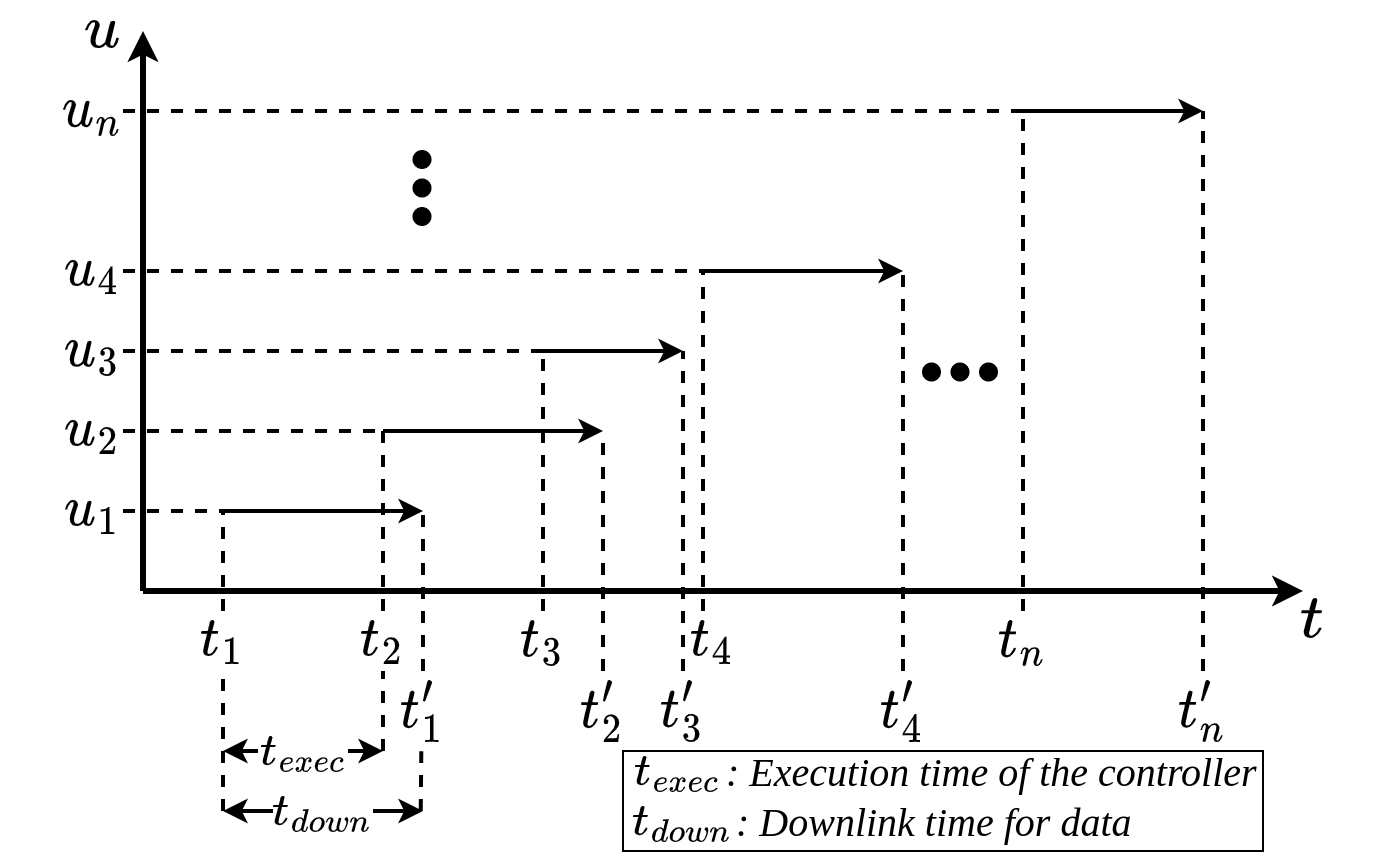}
  	\caption{Latency plot based on downlink time ($l_{c}$ and $l_{d}$)}
  	\label{fig:latency_plot}
\end{figure}

In Fig.~\ref{fig:latency_plot} the downlink latency is depicted, where $u_{i}$ is describing the command that is sent from the edge at time $t_{i}$ and arriving at the UAV at time $t_{i}'$. The frequency of the generated commands is set and considered fixed with the MPC rate denoted as $f_{exec}$ and the MPC execution time $t_{exec}$, constants. Thus, by measuring the time that a control command was generated and the time it arrived at the UAV, we can calculate the downlink latency. With this information and the measurements of the UAV's velocity, based on the last control command that was sent from the remote controller (MPC), we can estimate the position error, $\hat{e}$. Hence, the estimated error $\hat{e}_{i}$ is calculated by the distance, $d = v(t_{i-(k+1)}) \cdot l(t_{i})$ the UAV covered between two consecutive commands ($u_{i}$ and $u_{i-(k+1)}$).

During this work, two ways of downlink latency formulations and, respectively, two ways of error estimations are used to capture the UAV's behavior. The first estimated error $\hat{e}_{c}(t_{i})$, is based on the latency between two consecutive commands ($u_{i}$, $u_{i-(k+1)}$) that arrive to the UAV and is described by Eq.~\ref{eq:latency1}.

\begin{align}
    & \hat{e}_{c}(t_{i}) = v(t_{i-(k+1)}) \cdot l_{c}(t_{i}) \nonumber\\
    & l_{c}(t_{i}) = t_{i}' - t_{i-(k+1)}'  \label{eq:latency1}
\end{align}
where $k$ is the number of dropped packets, $v(t_{i-(k+1)})$ is the velocity of the UAV, based on the previous valid command (the command that was generated from the correct corresponding states of the UAV), and $l_{c}(t_{i})$ is the latency between two consecutive commands at the UAV ($u_{i}$ and $u_{i-(k+1)}$), for time stamps $t_{i}, i=1,2,..,n$.

The second estimated error $\hat{e}_{d}(t_{i})$ is based on the latency that is introduced by the time the command ($u_{i-k}$) was created on the edge server to the time the command ($u_{i}$) reached the UAV, and is described by Eq.~\ref{eq:latency2}.

\begin{align}
    & \hat{e}_{d}(t_{i}) = v(t_{i-(k+1)}) \cdot l_{d}(t_{i}) \nonumber\\
    & l_{d}(t_{i}) = t_{i}' - t_{i} + t_{exec} \cdot k  \label{eq:latency2}
\end{align}
where $l_{d}(t_{i})$ is the latency that is introduced by the time the command $u_{i-k}$ was generated by the edge and the time the valid command ($u_{i}$) arrived to the UAV.

When $t_{down} = t_{exec}$ then $\hat{e}_{c} = \hat{e}_{d}$, while when $t_{down} > t_{exec}$ then $\hat{e}_{c} < \hat{e}_{d}$ and $t_{down} < t_{exec}$ then $\hat{e}_{c} > \hat{e}_{d}$. Thus, to estimate the error $\hat{e}$ that is introduced to the system due to the latency, we calculate the mean error between $\hat{e}_{c}$ and $\hat{e}_{d}$, Eq.~\ref{eq:error}.

\begin{align}
    &\hat{e} = \frac{\hat{e}_{c}(t_{i}) + \hat{e}_{d}(t_{i})}{2} \label{eq:error}
\end{align}

Once we have estimated the error $\hat{e}$ that the latency can add to the system, we can propose a threshold $e_{th}$ to the accepted error.

\begin{figure*}[t]
	\centering
	\includegraphics[width=0.88\textwidth]{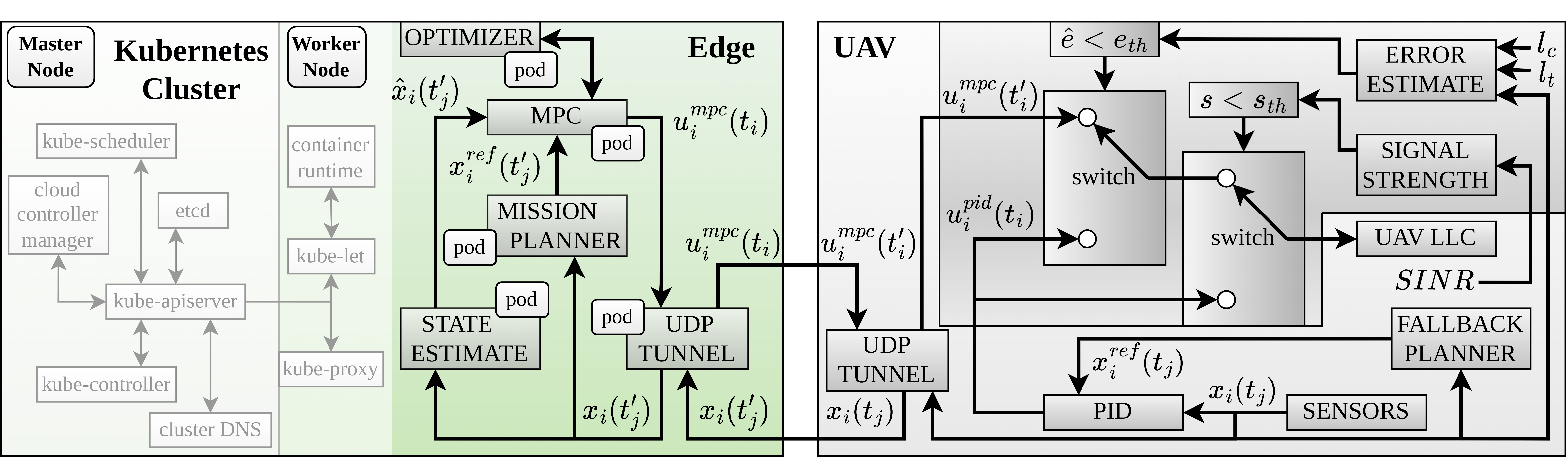}
  	\caption{Overall block diagram of the system with all the modules. The switching mechanism is highlighted with all the described components}
  	\label{fig:overall_block_diagram}
\end{figure*}

\subsubsection{Signal Strength}
\label{signal_strength}

In addition, other KPIs that can inform us about the status of the communication are the ones that describe channel conditions. Even though in some cases the latency might be low, there is the possibility of suddenly losing communication. The signal strength and more specifically the Signal-to-Interference-plus-Noise Ratio ($SINR$) can alert the system before the communication is lost and trigger the switch to activate the onboard control mode. A threshold $s^{th}$ based on $SINR$ studies have been utilized and set so that the latency and throughput requirements for the offloaded processes are met. 

\subsubsection{Switch Formulation}
\label{switch_formulation}

The system requirement is to keep the position error bounded and to ensure a reliable communication channel. It is impossible to eliminate the error completely because the system itself introduces some error as mentioned above. Our goal is to keep the error bounded in acceptable values for the safety of the UAV. Since we know that a portion of the error depends on the end-to-end latency of the system as well as the dropped packet count in the application layer, then we can estimate whether the latency error $\hat{e}$ will exceed a threshold $e_{th}$ that has been ad-hoc defined. Thus, we can predict that the overall error ($|p_{i} - p_{i}^{ref}|$) will exceed a predefined desired bounded limit (defined by use case characteristics), and then the switch will be triggered and turn the system into onboard control mode. Once the latency is low and the estimated error $\hat{e}$ is less than the $e_{th}$ ($\hat{e} < e_{th}$), the switch will be triggered and set the system back to the offboard control mode. Though the onboard-PID controller has the worst performance and the position error is overall bigger than the offboard-MPC, still, for safety reasons, the operation of the system using the PID controller is better and preferable when the latency is high.

In series to the error switch, a $SINR$ switch is placed as depicted in Fig.~\ref{fig:overall_block_diagram}. This switch is triggered when the signal $s$ based on the $SINR$ metrics exceeds the $s^{th}$. This switch is connected directly to the LLC of the UAV, thus, the $SINR$ can switch the system to the onboard control mode regardless of the output of the error switch.

To avoid undesired continuous changes in the control mode, a sliding window has been utilized. The switching strategy can be characterized as a two-level switch (one error switch and one $SINR$ switch). The overall system is shown in Fig.~\ref{fig:overall_block_diagram} with all the described components and modules.

\section{Experimental Evaluation}
\label{experiments}
For the experimental components, the following equipment was utilized. A real-life 5G network operating in mid-band frequencies (3.7 GHz) at the premises of the Luleå University of Technology. The used 5G network system provides an indoor 5G Ericsson DOT base station system that was chosen for the manifestation of the corresponding experiments. The utilized edge server is located near the local core breakout of the 5G network, thus achieving optimal low latency, which has been thoroughly demonstrated and documented in previous works \cite{damigos2023towards}. Further, a Kubernetes-enabled subset of the available resources was used regarding the edge server component. Finally, the used UAV is a Crazyflie 2 model, which is assisted by a Vicon motion capture system that captures the robot's states and further provides ground truth accuracy. 

To demonstrate the switching solution, network traffic that exceeds the UE's uplink capabilities was initiated by the 5G-edge-enabled UAV itself. This approach demonstrates realistic scenarios where various design components may affect the UAV's performance. The reader can find a comprehensive explanation of such scenarios in \cite{damigos2023performance}.

A characteristic mission designed to test an essential component required in most full-scale UAV missions was considered to evaluate the proposed architecture. The UAV takes off and has to execute a circular trajectory. The trajectory following problem is a commonly applicable building block to most complex UAV missions. The UAV executes the circular mission while offloading all the required processing to the edge server. During the mission, the system is disturbed, so the connectivity conditions deteriorate. More specifically, data that severely exceed the uplink capabilities of the 5G-edge-enabled UAV are striving to transmit, thus, significantly affecting the latency performance of the 5G-edge-enabled UAV system. Further, in a separate experiment, severe signal interference is induced at consecutive times. The architecture is tested in the aforesaid scenarios.

\begin{figure}[ht!]
	\centering
	\includegraphics[width=0.96\columnwidth]{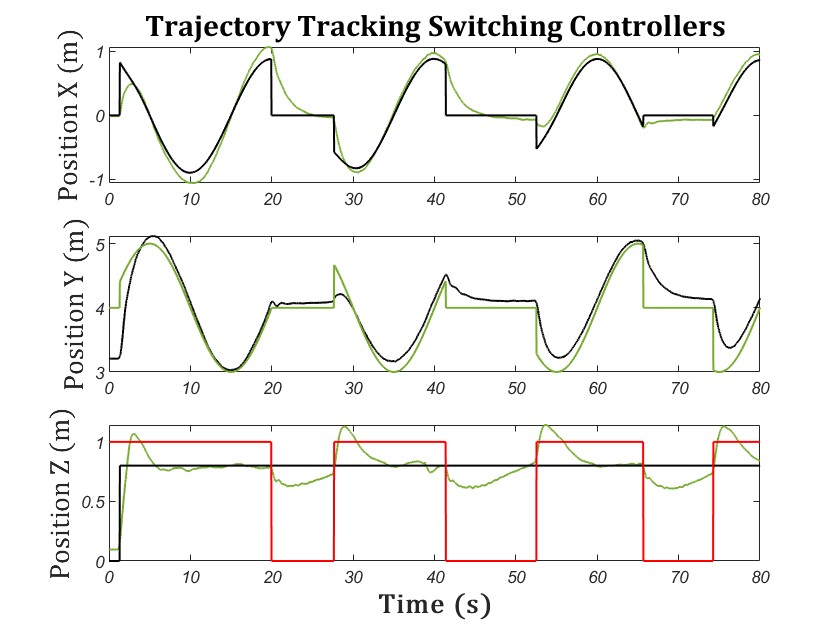}
  	\caption{UAV trajectory in 3 axes. The switching functionality to the onboard PID controller is triggered in challenging connectivity conditions. The UAV initially operates with the 5G-edge-enabled MPC controller. During this experiment, three switches are observed. The UAV alternates between estimated \textit{“safe” and “unsafe”} conditions. The switching decision is depicted with the red color signal.}
  	\label{fig:res-trajectories}
\end{figure}

\begin{figure}[ht!]
	\centering
	\includegraphics[width=0.96\columnwidth]{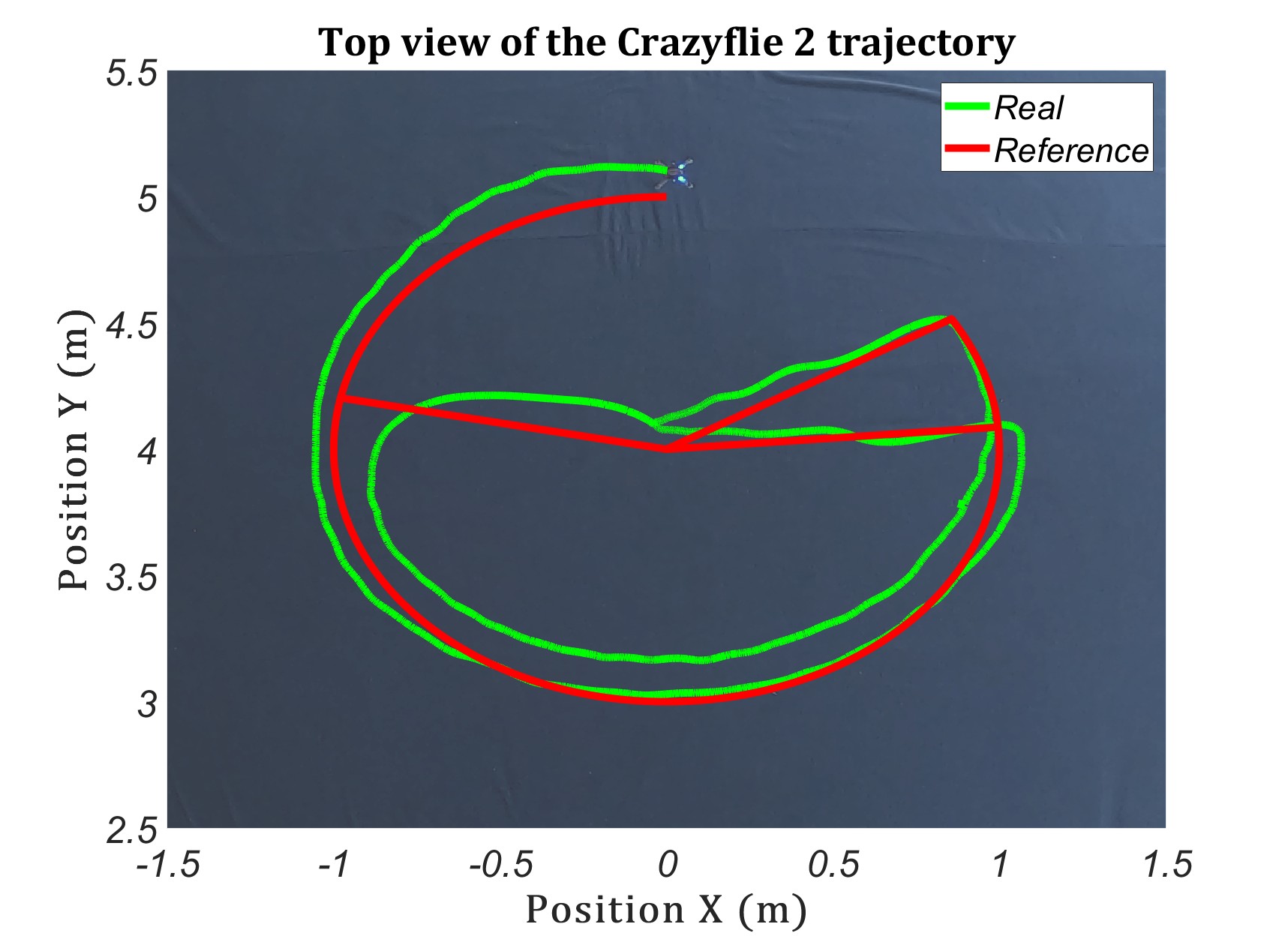}
  	\caption{Top view of the real and reference trajectories of the Crazyflie. The Crazyflie is commanded to do a circular trajectory when controlled by the offboard-MPC and is commanded to go to the home position $4p_{home} = [0, 4, 0.8]$ (center of the circle) and hover when controlled by the onboard-PID}
  	\label{fig:top_view}
\end{figure}

Fig. \ref{fig:res-trajectories} depicts the system's behavior when it is examined for different $SINR$ values, and Fig. \ref{fig:top_view} depicts the top view of the described experiment. Each row shows the reference or “desired” trajectory and the real or captured trajectory. Low $SINR$ values trigger the switch at a selected threshold of $s^{th} \,= \, 6 \,dB$. The additional interference that causes the $SINR$ drop is enabled and disabled three times. This experiment seeks to validate the system's behavior and the switching controllers' performance. Additionally, the chosen threshold expresses frequent scenarios, such as near-cell edge conditions or conditions where the UAV experiences severe interference, e.g., high altitude flights \cite{zeng2019accessing}. Note that the data rate requirements for transmitting the robot's state, i.e., the captured states by the Vicon system and the controlled commands sent by the edge server to the UAV, require small data rates compared to the available capabilities. For quantification purposes, the UAV's states require $\sim 30 Kbps$, the control commands require $\sim 15 Kbps$, and the uplink and downlink capabilities of the considered system are $\sim 94 \, Mbps$ and $\sim 1402 \, Mbps$ respectively. Consequently, the system's latency is not linearly correlated to the $SINR$ values. However, if the channel conditions deteriorate, this would yield a selection of a lower modulation scheme and consequently lower the achievable throughput, for example, 64 QAM (Quadrature Amplitude Modulation) would be selected in relatively good channel conditions, whereas 16 QAM would be selected in worsen channel conditions. In conclusion, when the channel conditions deteriorate enough to enable a modulation scheme that considering the remaining combined traffic would not be able to comprehend the overall uplink or downlink transmission, then latency rise will be observed in the control and command packets as well as in the robot's state packets. Additionally, if the $SINR$ values drop significantly enough, the system's connectivity is completely discontinued. Finally, it is important to note that the system's tracking accuracy decreases when the PID controller takes over, and transient effects on the switching state of the two controllers are visible. Such challenges can be addressed with extensive tuning, additional controllers that target the transient phase, and others; nevertheless, it is not within the scope of this work. Overall, the proposed switching strategy demonstrated successfully that the system is able to fallback into the onboard controller when the $SINR$ values are below the chosen threshold. 

\begin{figure}[ht!]
	\centering
	\includegraphics[width=0.96\columnwidth]{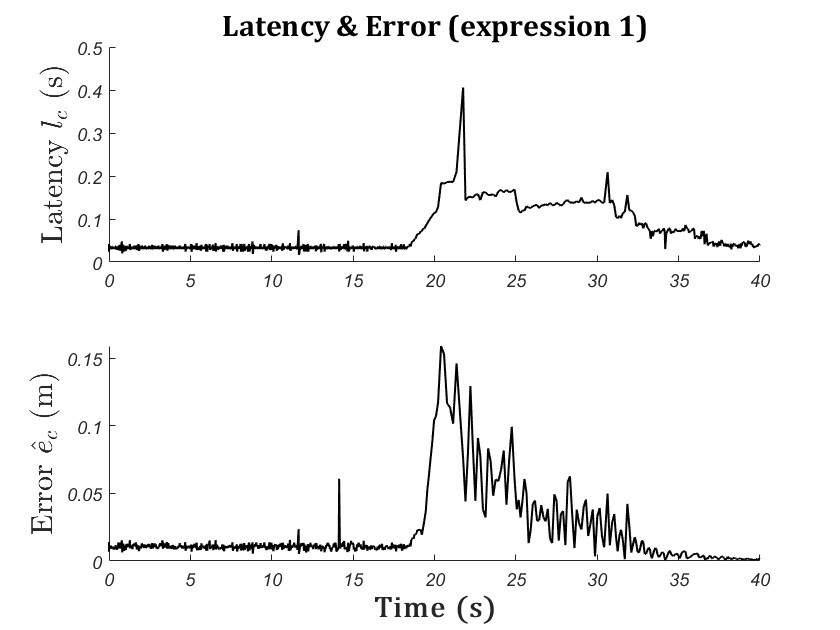}
  	\caption{Estimated error $\hat{e}_{c}$ along with the corresponding measured latency $l_c$. Note that the induced latency is initiated at the $\sim 18.5 \, s$.}
  	\label{fig:res1}
\end{figure}

\begin{figure}[ht!]
	\centering
	\includegraphics[width=0.96\columnwidth]{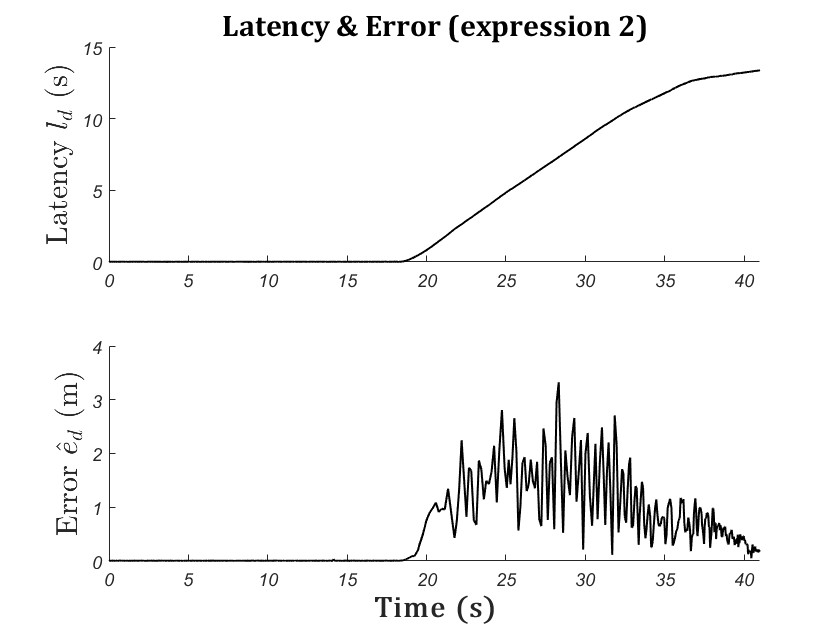}
  	\caption{Estimated error $\hat{e}_{d}$ along with the corresponding measured latency $l_d$. Note that the induced latency is initiated at the $\sim 18.5 \, s$.}
  	\label{fig:res2}
\end{figure}

Regarding the latency aspect of the proposed safety mechanism, a situation where the UAV was performing the autonomous mission of executing a circular trajectory on a set height was examined. During this scenario, a remote MPC controller was operating in a K8s pod hosted on the edge server. The communication of the UAV and the edge server was established over a 5G network. Subsequently, to test the performance of the switching mechanism, an attempt to transmit sensor data that significantly surpassed the 5G-edge-enabled UAV's uplink capabilities was initiated; hence increased latency was inducted into the end-to-end system's data packet flows. Similar real-life scenarios commonly occur when the design process regarding the participating data flow profiles fails. A common factor that could produce such scenarios revolves around the unique characteristics of applications that exhibit large fluctuations in the produced data rates (e.g., image processing algorithms). Another one relates to the unique character of UAV communications. For example, the mobility of the UAV might strongly affect the communication channel conditions. The combination of the two latter paradigms can create scenarios that, without a large enough safety margin, the requested data rates on the UAV side might exceed the communication system's capabilities. 

Fig. \ref{fig:res1} depicts the measured latency $l_c$ and the corresponding estimated error $\hat{e}_c$, while Fig. \ref{fig:res2} depicts the measured latency $l_d$ and the corresponding error $\hat{e}_{d}$. In both figures, it is visible that the latency was initiated approximately at the $18.5 \, s$ of the mission. The increase in both latency formulations is directly observed, which is also captured in the corresponding errors. However, each formulation presents different sensitivity and, thus, different error estimations. For example, it is evident that the dropped packet rate significantly affects the $\hat{e}_{d}$. Another contributing factor to the large fluctuations is the velocity of the UAV, which also presents large fluctuations. This is mostly an outcome of the downlink delay, which causes the UAV to be in incorrect positions, then the UAV tries to address its incorrect position aggressively. For those reasons, even though both formulations capture the expected error, the combined average metric is preferred for the initiation of the switching action.    

\begin{figure}[ht!]
	\centering
	\includegraphics[width=0.96\columnwidth]{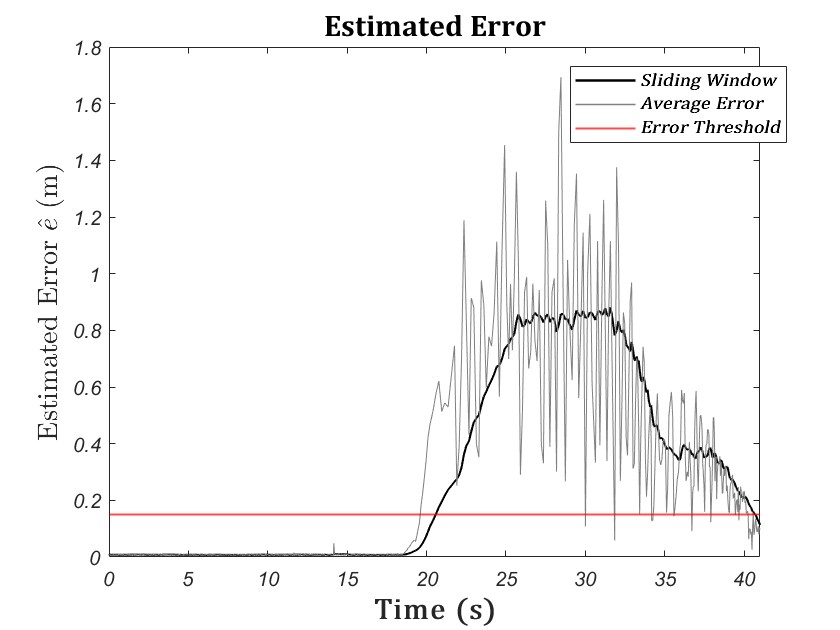}
`````````  	\caption{Estimated error of the UAV. The gray curve demonstrates the UAV estimated raw error time series with the corresponding high fluctuating frequency. The black curve demonstrates the UAV's sliding window error estimation, which is utilized for the switching functionality. The red curve depicts the error threshold or the switching condition. Here, the error threshold is set to the value of 0.15 m.}
  	\label{fig:res3}
\end{figure}

The combined error $\hat{e}$ is depicted in Fig. \ref{fig:res3}. This formulation is used to calculate the switching condition. It is evident that the $\hat{e}$ inherits the strong fluctuation characteristic of $\hat{e}_{c}$ and $\hat{e}_{d}$. This fact would deem this metric unstable to be directly used. The metric would yield multiple switches within the duration of the increased latency and thus would risk the system's stability. To address that, as mentioned in Section \ref{switch_formulation}, a sliding window average is applied in $\hat{e}$. Both the $\hat{e}$ and the corresponding sliding window formulation is depicted in Fig \ref{fig:res3}. For this experiment, the window size is set to 50 samples, and the error threshold $e_{th} = 0.15 \, m$. Please note that the estimated error of 0.15 m refers only to the error produced by the latency effect. The system identifies the expected error and switches to the onboard PID controller. Then, when the latency is disabled and the sliding window average error becomes smaller than the decided threshold, the UAV switches back to the optimal remote controller. Overall, the validation of the system performed as expected, and the switching strategy ensured the system's stability. 

\section{Conclusions and Future Work}
\label{conclusions}
In this article, a novel switching strategy was presented. This strategy ensures that a UAV will operate in one of the two following modes, based on a resilient reactive mechanism that uses the available communication KPIs. The first mode is the offboard mode (edge-based mode), which utilized an MPC for controlling the trajectory of the UAV. The MPC has been offloaded to the edge for an optimized performance and the communication is over 5G. The offboard mode is active as long as the switching mechanism does not detect any issue on the communication channel. Once the channel is considered non-reliable, or the communication link is unstable, the switching mechanism turns the system to the onboard mode for safety and redundancy reasons. The system stays in onboard safety mode as long as required. Once the metrics indicate that the channel is reliable again, the system turns to the offboard mode. The validation of the proposed switching strategy was thoroughly tested in laboratory experiments.

The field of edge robotics has room for many different directions. Some related interesting future implementations could be the investigation of mobile edge computing for migrating from one edge cluster to another based on the communication, or the development of algorithms to ensure the safe, rapid, and reliable redeployment of the mission-critical application at the edge through a Kubernetes cluster. Finally, an interesting study would be the task allocation, management, and control of multiple collaborative robots through the edge for real-time applications.

\bibliographystyle{./IEEEtranBST/IEEEtran}
\bibliography{./IEEEtranBST/IEEEabrv, references}

\end{document}